\documentclass[conference]{IEEEtran}
\IEEEoverridecommandlockouts
\usepackage[utf8]{inputenc}
\usepackage{cite}
\usepackage{amsmath,amssymb,amsfonts}
\usepackage{algorithmic}
\usepackage{graphicx}
\usepackage{textcomp}
\usepackage{xcolor}
\usepackage{comment}
\usepackage{float}
\usepackage{afterpage}
\usepackage{placeins}      % provides \FloatBarrier
\usepackage{stfloats}      % improves two-column float 
\usepackage{multirow}

\def\BibTeX{{\rm B\kern-.05em{\sc i\kern-.025em b}\kern-.08em
    T\kern-.1667em\lower.7ex\hbox{E}\kern-.125emX}}
    
\begin{document}

\title{Contactless Respiratory Monitoring on Heterogeneous Mobile Robots: A Multimodal Edge-Computing Framework\\
}

% \author{
%   \IEEEauthorblockN{
%      Milind Rampure\IEEEauthorrefmark{1},
%      Shadman Sakib\IEEEauthorrefmark{1},
%      Haley Patel\IEEEauthorrefmark{2},
%      Zahid Hasan\IEEEauthorrefmark{1},
%      Nirmalya Roy\IEEEauthorrefmark{1},
%    }
%    \IEEEauthorblockA{
%      \IEEEauthorrefmark{1}Department of Information Systems, University of Maryland Baltimore County, USA
%    }
%    \IEEEauthorblockA{
%      \IEEEauthorrefmark{2}Department of Computer Science,
%      University of Maryland Baltimore County, USA
%    }
%  }
% \maketitle

\author{
  \IEEEauthorblockN{
     Milind Rampure\textsuperscript{*}\textsuperscript{\textdagger},
     Shadman Sakib\textsuperscript{*}\textsuperscript{\textdagger},
     Haley Patel\textsuperscript{\textdaggerdbl},
     Zahid Hasan\textsuperscript{*},
     Nirmalya Roy\textsuperscript{*}
  }

  \IEEEauthorblockA{
     \textsuperscript{*}Department of Information Systems,
     University of Maryland Baltimore County, USA
  }

  \IEEEauthorblockA{
     \textsuperscript{\textdaggerdbl}Department of Computer Science,
     University of Maryland Baltimore County, USA
  }

  \thanks{\textsuperscript{\textdagger}Equal contribution.}
}

\maketitle
\begin{abstract}
Respiratory-rate (RR) monitoring is a critical component of remote triage and victim assessment in emergency response, disaster recovery, and infectious-disease scenarios, where minimizing physical contact can reduce responder risk and improve operational safety. However, field deployment of contactless RR monitoring remains challenging due to variable illumination, posture changes, platform heterogeneity, and the impracticality of wearable sensors in hazardous environments. In this paper, we present a modality-adaptive contactless RR monitoring framework for heterogeneous mobile robots with onboard edge computing. The proposed system combines brightness-adaptive sensor selection across RGB, thermal, near-infrared (NIR), and low-light cameras, keypoint-guided chest ROI extraction for posture-robust monitoring, and a signal-quality-index (SQI)-based filtering mechanism for reliable respiratory estimation. We implement and evaluate the framework on three robotic platforms spanning quadruped and wheeled locomotion and multiple edge-computing architectures. Experiments conducted across diverse lighting conditions, subject poses, and robot-to-subject distances demonstrate that the framework generalizes across platforms without per-platform algorithmic retuning, while revealing modality-specific operational boundaries. RGB provides the broadest coverage up to 8\,m, NIR remains effective up to 6\,m, thermal is reliable only at short range, and low-light sensing supports monitoring in complete darkness up to 8\,m. Overall, the results demonstrate the feasibility of multimodal contactless RR monitoring on mobile robots and support its use as a foundation for autonomous triage and victim assessment in hazardous search-and-rescue settings.
\end{abstract}


\section{Introduction}

Respiratory rate (RR), defined as the number of breaths per minute, is a critical vital sign for assessing physiological status and detecting respiratory distress, clinical deterioration, and other life-threatening conditions \cite{majumder2017wearable}. It also reflects stress, fatigue, physical exertion, and cognitive workload \cite{massaroni2020contactless, sakib2025state}. However, RR remains one of the least consistently monitored vital signs because contact-based methods, such as chest straps, nasal probes, and wearable sensors, are often uncomfortable, restrictive, and impractical in remote or dynamic environments \cite{massaroni2019contact}. \begin{figure}[!ht]
    \centering   \includegraphics[width=\columnwidth]{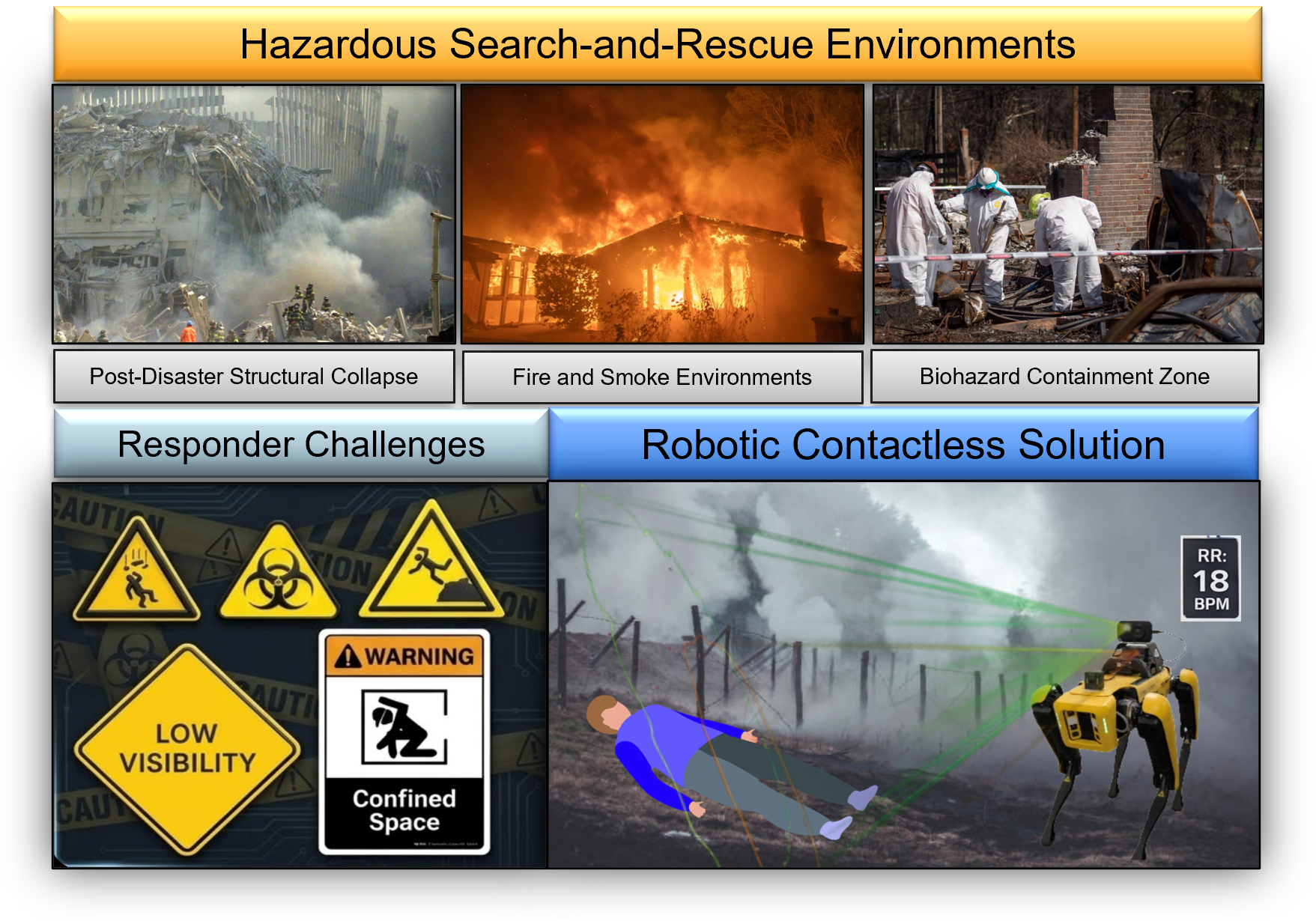}
    \caption{Adaptive multi-modal respiratory-rate (RR) monitoring on a mobile robotic platform under variable illumination and environmental conditions.}
    \label{fig:overview}
\end{figure}To address these limitations, contactless RR monitoring has gained increasing attention. Existing approaches use video, radar, and thermal imaging to capture subtle respiration-induced motion \cite{massaroni2019comparison,sakib2025respformer}. Camera-based methods are particularly attractive because they are low cost, widely available and suitable for scalable deployment in clinical and emergency settings \cite{queiroz2020video, darpaTriagChallenge}. Earlier work relied on classical signal-processing techniques such as optical flow, motion magnification, pixel-difference analysis, and band-pass filtering \cite{manne2023automatic, massaroni2019contact}, while more recent studies use deep learning with RGB, NIR, and thermal data \cite{sakib2025e2respunet}. Despite these advances, robust field deployment remains difficult because performance is still affected by illumination changes, body motion, weak cross-subject generalization, and computational cost \cite{massaroni2019comparison, majumder2017wearable}. These challenges are amplified in emergency and disaster-response scenarios, where rapid and contactless assessment of multiple casualties is needed while limiting first-responder exposure to hazardous environments \cite{darpaTriagChallenge, cruz2021autonomous}. In such settings, stationary systems are often impractical, and single-modality approaches can fail as conditions change. For example, RGB sensing may work well under favorable lighting but can degrade sharply in darkness, smoke, or visually degraded scenes \cite{alnaggar2023video}. As illustrated in Figure~\ref{fig:overview}, these environments require sensing strategies that can adapt to changing visibility and operational constraints.

Mobile robots provide a promising platform for this problem because they can access rubble, uneven terrain, and confined spaces that may be unsafe for human responders \cite{hoeller2024anymal}. However, deploying contactless RR monitoring on robots introduces additional challenges, including variable subject-to-camera distance, changing illumination, heterogeneous sensor payloads, and limited onboard computing resources \cite{huang2022mobile, cruz2023mixed}. Although prior work such as Dr. Spot has shown the feasibility of thermal-RGB sensing on quadruped robots \cite{huang2022mobile, damavsevivcius2023sensors}, systematic evaluation across heterogeneous robotic platforms with different sensing and edge-computing configurations remains limited.

In this paper, we present a \textbf{modality-adaptive contactless respiratory monitoring} framework for heterogeneous mobile robots with onboard edge computing. The proposed system combines brightness-adaptive sensor selection across RGB, thermal, near-infrared (NIR), and low-light cameras, keypoint-guided chest ROI extraction for posture-robust monitoring, and a signal-quality-index (SQI)-based filtering strategy for reliable respiratory-rate (RR) estimation under variable field conditions. The framework is evaluated on three mobile robotic platforms spanning quadruped and wheeled locomotion, diverse sensing payloads, and distinct edge-computing architectures. By integrating multimodal sensing, robotic mobility, and real-time onboard processing, this work supports autonomous triage and victim assessment in hazardous environments where conventional contact-based or stationary monitoring is not feasible.

The primary contributions of this paper are as follows:
\begin{itemize}
    \item A modality-adaptive contactless respiratory monitoring framework for heterogeneous mobile robots that supports RGB, thermal, NIR, and low-light sensing modalities.

    \item An edge-computing pipeline that combines brightness-adaptive modality selection, keypoint-guided chest ROI extraction, respiratory signal processing, and SQI-based window filtering for robust onboard RR estimation.

    \item A systematic evaluation across three robotic platforms, multiple environmental conditions, subject poses, and stand-off distances, establishing modality-specific operational envelopes and demonstrating utility for autonomous triage and victim assessment in hazardous search-and-rescue environments.
\end{itemize}

\section{Related Works}
% \vspace{-0.1in}
%While previous works at AI-RMH 2024 explored resilience in wearable devices [Citation], our work extends this to autonomous mobile agents by introducing a brightness-adaptive modality selection logic.”
%“Unlike [Citation], which focuses on stationary RGB sensing, our framework utilizes modality-specific SQI thresholds to allow for the higher noise floor inherent in thermal and low-light sensors on moving robotic platforms.

Contactless respiratory-rate (RR) estimation from video has progressed from early proof-of-concept studies to increasingly realistic monitoring scenarios. Early methods primarily inferred breathing from subtle torso or chest motion using frame differencing \cite{tan2010real}, optical flow \cite{queiroz2020video,janssen2015video}, and phase-based analysis \cite{massaroni2019contact,tveit2016motion}. These studies established the feasibility of video-based RR estimation, but performance often degraded under subject motion, posture variation, occlusion, and illumination changes. Later work improved robustness through more structured motion modeling, including head-movement tracking to mitigate occlusion-related errors \cite{Gwak2023} and artifact-removal methods to stabilize respiratory signal extraction under moderate motion \cite{Gwak2022MotionRespiratoryRate}. More broadly, foundational non-contact vital-sign monitoring work showed that physiological signals can be recovered from ambient-light video \cite{tarassenko2014non}, while later survey, real-time monitoring, and clinical validation studies reinforced the potential of contactless monitoring in both laboratory and medical settings \cite{molinaro2022contactless,alnaggar2023video,goldfine2024contactless}.

Recent work has also explored thermal and multimodal imaging to address the limitations of RGB-only sensing. Thermal video remains effective in low-light or nighttime conditions where RGB methods degrade, whereas RGB typically provides higher spatial detail under favorable illumination. Comparative studies highlight this complementary trade-off and show that neither modality is uniformly reliable across operating conditions \cite{yang2022contactless}. Related work on remote photoplethysmography under real-world and extreme lighting conditions further underscores the sensitivity of camera-based physiological estimation to illumination changes outside controlled environments \cite{shao2025remote, DiLernia2024rPPG}. Together, these results motivate modality-aware sensing and multimodal adaptation for deployment scenarios in which viewpoint, subject distance, and ambient lighting cannot be tightly controlled.

% Signal quality assessment is another key requirement for reliable physiological inference. Prior studies introduced signal-quality indices based on template matching, spectral consistency, autocorrelation, and related measures to distinguish valid from corrupted physiological measurements \cite{elgendi2024optimal,song2019pqr}. Quality-aware video methods have likewise shown that explicit reliability modeling can improve respiratory estimation and waveform reconstruction from image sequences \cite{hasan2025rrpips}. Contactless sensing has also begun to move onto robotic platforms. 
Signal quality assessment is essential for reliable physiological inference. Prior work has proposed signal-quality indices based on spectral consistency, autocorrelation, and related measures to distinguish valid from corrupted physiological measurements \cite{elgendi2024optimal,song2019pqr}. Quality-aware video methods have similarly shown that explicit reliability modeling can improve respiratory estimation and waveform reconstruction \cite{hasan2025rrpips}. More recently, contactless sensing has begun to extend to robotic platforms. A mobile robotic platform for contactless vital-sign monitoring was demonstrated in \cite{huang2022mobile}, while related search-and-rescue and emergency-response systems have focused more broadly on victim detection, robotic triage, and field operation in complex environments \cite{cruz2021autonomous,senthilkumaran2023artemis,cruz2023mixed}. Advances in quadruped locomotion, heterogeneous robot-team deployment, emergency sensing infrastructures, and edge intelligence further support deployment in challenging real-world settings \cite{hoeller2024anymal,damavsevivcius2023sensors,tranzatto2022team}. However, prior work has not systematically examined contactless RR monitoring across heterogeneous robotic platforms with non-uniform sensor payloads, diverse edge-computing constraints, and modality-specific quality adaptation requirements. This gap motivates the present work.

\section{Methodology}
\begin{figure*}[!t]
  \centering
\includegraphics[width=\textwidth]{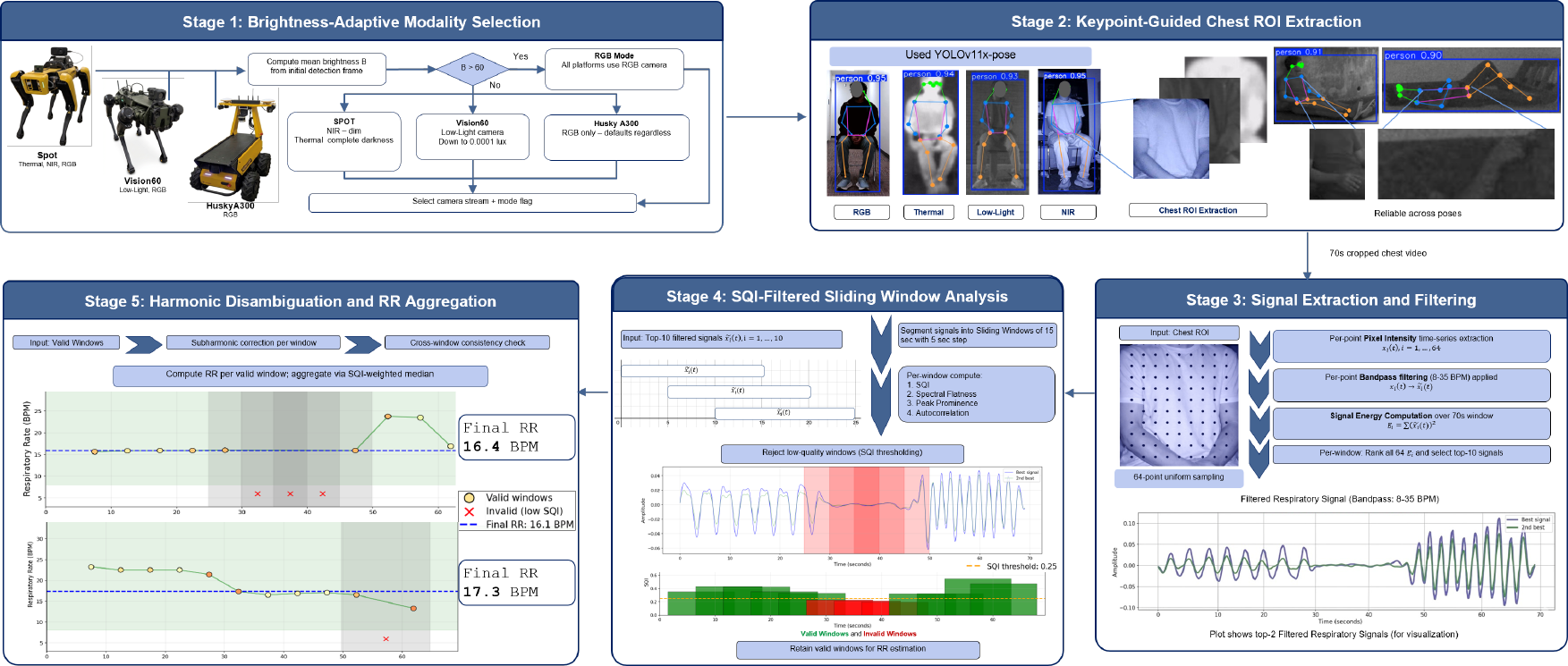}
  \caption{System methodology workflow illustrating detection, ROI extraction, temporal signal processing, and RR estimation pipeline.}
  \label{fig:method}
\end{figure*}

We propose a modality-adaptive respiratory-rate (RR) estimation pipeline that can be deployed across heterogeneous robotic platforms without requiring platform-specific algorithmic redesign. As shown in Figure~\ref{fig:method}, the pipeline comprises five stages: (i) platform-specific video acquisition and adaptive modality selection, (ii) 2D keypoint-guided chest ROI extraction, (iii) respiratory signal extraction and filtering, (iv) SQI-filtered sliding-window analysis, and (v) harmonic disambiguation with final RR aggregation.

\subsection{Video Acquisition and Sensor Selection}
Each robotic platform acquires video through its native interface, including USB-connected sensors, ROS\,2 image topics, and forwards the resulting frames to a common processing pipeline. This abstraction decouples the downstream RR estimation stages from platform-specific sensing and communication details while enabling a unified algorithmic workflow across the heterogeneous systems summarized in Table~\ref{tab:platforms}. A brightness-adaptive sensor-selection module determines the operating modality from the mean brightness $B$ of the initial detection frame. For all platforms, RGB sensing is selected when $B > 60$. When $B \leq 60$, the framework switches to an alternative modality according to the sensing payload available on the robot. On SPOT, NIR is preferred in dim scenes because active 850\,nm illumination remains effective up to 6\,m (Table~\ref{tab:distance_bounds}),whereas thermal sensing is selected under near-complete darkness. On Vision~60, the low-light camera is used for all sub-threshold conditions. On Husky~A300, which carries only an RGB sensor, the pipeline defaults to the available modality regardless of brightness. This adaptive selection strategy allows the same downstream processing stages to operate robustly under variable illumination and sensing availability.

\subsection{2D Pose-Based Chest ROI Extraction}
Person localization and 2D pose estimation are performed using YOLOv11x-pose across all sensing modalities. Fixed proportional crops derived directly from the person bounding box are highly sensitive to posture and often fail to remain aligned with the thoracic region when the subject is sitting, leaning, or lying down. To address this limitation, we derive the chest ROI directly from the detected left/right shoulder and left/right hip keypoints. This keypoint-guided formulation yields a geometry-adaptive crop that remains aligned with the thorax across posture variation and is more tolerant to partial keypoint occlusion. We selected YOLOv11x-pose over the lighter YOLOv11n-pose variant because the latter produced lower keypoint confidence and occasional detection failures at extended distances ($>6$\,m), particularly in non-upright poses and under thermal and low-light imaging. After ROI localization, a 70-second chest-cropped video segment is forwarded to the temporal processing stage.

\subsection{Signal Extraction and Filtering}
To avoid transient exposure adjustments, the first 30 frames of each cropped video segment are discarded. Time-series signals are then extracted from $N{=}64$ grid points distributed uniformly across the chest ROI. At each grid point, a $20{\times}20$ pixel block is spatially averaged in every frame to produce one temporal intensity signal $x_i(t)$, where $i{=}1,\ldots,64$. The signal source depends on the active sensing modality. For RGB video, the green channel is used because it provides higher sensitivity to subtle surface changes associated with thoracic expansion. For thermal and low-light imagery, which effectively provide grayscale outputs with identical RGB channels, the mean pixel intensity across the chest ROI is used. For NIR imagery, the mean intensity is also extracted because the observed pseudo-color variations result from differential sensor response to 850\,nm illumination rather than physiologically meaningful color content. Each signal is then bandpass-filtered using a fifth-order Butterworth filter with passband 8--35\,BPM (0.13--0.58\,Hz), preserving physiologically plausible respiratory frequencies while suppressing motion artifacts and low-frequency drift~\cite{molinaro2022contactless}. The filtered signals $\tilde{x}_i(t)$ are ranked by energy:
\begin{equation}
E_i = \sum_{t} \tilde{x}_i(t)^2,
\end{equation}
and the top-10 highest-energy signals are retained for downstream analysis, prioritizing grid locations that capture the strongest respiration-induced motion content.

\subsection{SQI-Filtered Sliding Window Analysis}
The top-10 filtered signals are analyzed using overlapping sliding windows of 15\,s with a 5\,s step. For each window, a three-component Signal Quality Index (SQI) is computed from the highest-energy signal. Before SQI estimation, each signal is normalized to zero mean and unit variance so that the quality measure reflects signal \emph{periodicity} rather than raw amplitude. Spectral flatness (Wiener entropy) measures the concentration of spectral energy within the respiratory band:
\begin{equation}
\mathrm{SF} =
\frac{\left(\prod_{k=1}^{K} S(k)\right)^{1/K}}
{\frac{1}{K}\sum_{k=1}^{K} S(k)},
\end{equation}
where $S(k)$ denotes the Welch power spectral density at frequency bin $k$ within the 8--35\,BPM band. Values near 1 indicate a flat, noise-like spectrum, whereas values near 0 indicate a peaked and periodic spectrum. Peak prominence is defined as the ratio of the dominant spectral peak power to the median power in the respiratory band, i.e., $P = S_{\mathrm{peak}} / \mathrm{median}(S)$. Autocorrelation confidence $R_{\mathrm{ac}}$ is defined as the normalized autocorrelation peak height at the lag corresponding to the dominant breathing period. These three terms are combined as a weighted sum:
\begin{equation}\label{eq:sqi}
\mathrm{SQI} = 0.3\,(1 - \mathrm{SF})
             + 0.4\,\min\!\left(1,\;\frac{P}{10}\right)
             + 0.3\,R_{\mathrm{ac}}.
\end{equation}

In addition to the continuous SQI score, each window is subject to hard rejection when either $\mathrm{SF}$ or $P$ violates its modality-specific threshold. In that case, the window is capped below the acceptance threshold regardless of the remaining SQI components.

The modality-specific SQI thresholds in Table~\ref{tab:sqi_thresholds} account for the distinct noise characteristics of the sensing modalities. Platform~A RGB, which exhibits the highest signal-to-noise ratio, uses the strictest thresholds. The Husky~A300 and Vision~60 RGB sensors require moderately relaxed criteria due to elevated sensor noise. NIR and thermal modalities use more lenient flatness thresholds to accommodate active-illumination limitations and lower spatial resolution, respectively. Windows that fail the modality-specific acceptance criteria are discarded from subsequent RR estimation.
\begin{table}[!ht]
\centering
\caption{Modality-Specific SQI Threshold Configuration}
\label{tab:sqi_thresholds}
\small
\setlength{\tabcolsep}{3pt}
\begin{tabular}{|l|ccccc|}
\hline
\textbf{Parameter} & \textbf{Plat. A} & \textbf{Plat. B\&C} & \textbf{NIR} & \textbf{Therm.} & \textbf{LL} \\
\hline
SQI $\geq$ & 0.25 & 0.25 & 0.25 & 0.20 & 0.25 \\
\hline
Flatness $<$ & 0.75 & 0.75 & 0.85 & 0.85 & 0.88 \\
\hline
Prominence $>$ & 1.5 & 2.0 & 1.5 & 1.5 & 1.5 \\
\hline
\end{tabular}
\end{table}

\subsection{Harmonic Disambiguation and RR Aggregation}
Within each valid window, respiratory rate is estimated using median voting across FFT-based and autocorrelation-based estimates computed from the top-10 retained signals. The resulting window-level estimate is then refined through harmonic disambiguation. If the subharmonic frequency at half of the dominant spectral peak carries more than 30\% of the dominant peak power, the estimate is corrected to the subharmonic value. The same procedure is applied to third-harmonic ambiguity at $f/3$. To improve temporal consistency, we further perform cross-window consistency checking. Windows whose RR estimates deviate from neighboring windows by approximately ${\approx}2{\times}$ or ${\approx}3{\times}$ are flagged as likely harmonic errors and corrected using the corresponding subharmonic when supported by the local spectral structure.

The final respiratory rate is computed as the SQI-weighted median across all accepted windows. Given $W$ valid windows with estimates $\{r_w\}$ and weights $\{\mathrm{SQI}_w\}$, the weighted median is defined as the value $r^*$ at which the cumulative weight of the sorted estimates reaches half of the total weight. If all windows are rejected, the pipeline reports that no reliable respiratory signal was detected.

\section{System Description and Experimental Setup}
We evaluate the proposed framework on three mobile robotic platforms that span two locomotion classes and three edge-computing architectures. As summarized in Table~\ref{tab:platforms}, Platform~A (SPOT) is equipped with three sensing modalities (RGB, thermal, and NIR), Platform~B (Vision~60) carries two modalities (RGB and low-light), and Platform~C (Husky~A300) operates with RGB only. This heterogeneous platform design enables us to assess how the proposed pipeline performs under varying sensing capabilities, mobility constraints, and onboard compute resources. \begin{table*}[!ht]
\centering
\caption{Heterogeneous Platform and Sensor Configuration}
\label{tab:platforms}
\begin{tabular}{lccc}
\hline
\textbf{Specification} &
\textbf{Platform A (Q-UGV)} &
\textbf{Platform B (Q-UGV)} &
\textbf{Platform C (Wheeled UGV)} \\
\hline
Base Platform        & Boston Dynamics SPOT        & Ghost Robotics Vision 60      & Clearpath Husky A300 \\
Locomotion           & Quadruped                   & Quadruped                     & Differential-drive wheeled \\
Edge Compute         & NVIDIA Jetson Orin AGX      & NVIDIA Jetson Xavier          & x86 + NVIDIA RTX 4060 Ti \\
GPU                  & 2048-core Ampere            & 512-core Volta                & Ada Lovelace (8 GB VRAM) \\
Memory               & 64 GB unified               & 32 GB unified                 & 32 GB RAM + 8 GB VRAM \\
Power Monitor        & tegrastats (VDD\_IN)        & jtop (Power TOT)              & nvidia-smi + RAPL \\
Modality count       & 3 (RGB, thermal, NIR)       & 2 (RGB, low-light)            & 1 (RGB only) \\
\hline
\multicolumn{4}{c}{\textbf{Sensing Modalities}} \\
\hline
RGB Camera           & Custom (1920×1080, 30 fps)  & Built-in (960×540, 20 fps)    & OAK-D Pro W (1334×1008, 15 fps) \\
Thermal Camera       & FLIR C5 (640×480, 10 fps)   & —                             & — \\
NIR Camera           & Custom w/ 6× 850nm LEDs     & —                             & — \\
                     & (1920×1080, 30 fps)         &                               & \\
Low-Light Camera     & —                           & Payload (960×540, 30 fps,     & — \\
                     &                             & 0.0001 lux min.)              & \\
\hline
\end{tabular}
\end{table*}Experiments were conducted across four environmental conditions: indoor lit (standard indoor illumination), indoor dark (lights off), outdoor lit (clear daytime conditions), and outdoor dark (nighttime with no artificial lighting). These scenarios were selected to reflect the diverse visibility conditions likely to arise in practical field deployment. For each scenario, trials were performed at robot-to-subject distances of 2, 4, 6, and 8\,m in order to evaluate the operational range of each sensing modality. Multiple subjects participated in the study to examine robustness across differences in body type and clothing. To further assess posture robustness, each experimental condition was evaluated using three subject poses: standing, seated, and lying. Every trial consisted of a 70-second chest video recording collected under a structured breathing protocol comprising three sequential phases: normal breathing (0--30\,s), breath-hold (30--50\,s), and deep breathing (50--70\,s). \begin{figure}[!ht]
    \centering
    \includegraphics[width=\columnwidth]{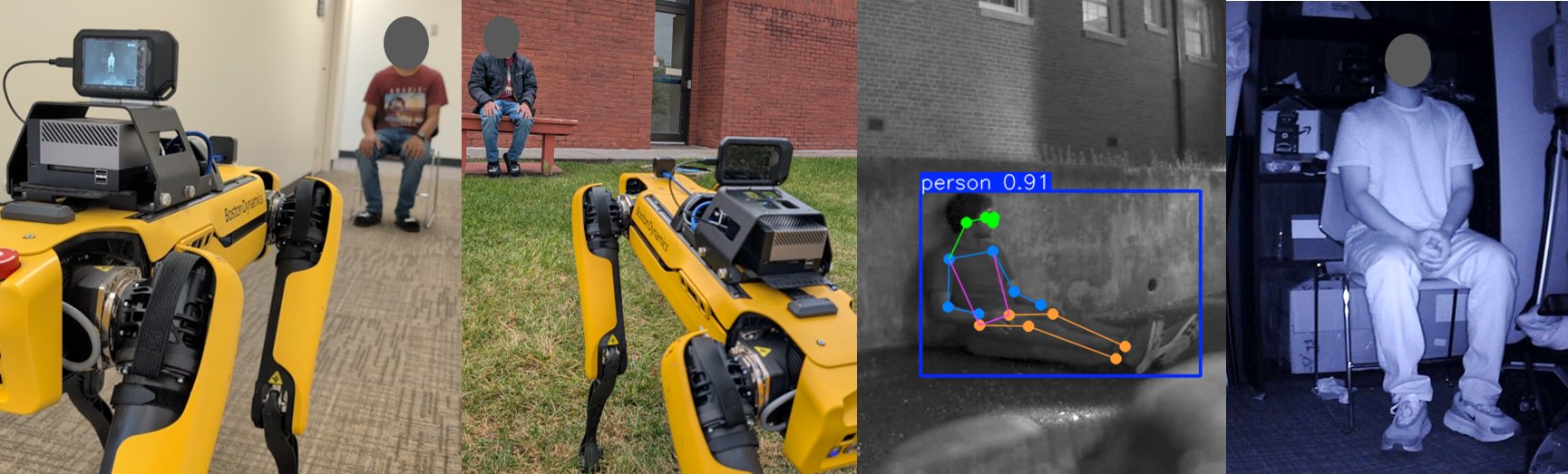}
    \caption{Representative experimental configurations: (left to right) Platform A with RGB camera at 2\,m indoor lit, Platform A at 4\,m outdoor lit, Platform B low-light capture of lying subject at 2\,m with YOLOv11x-pose keypoint overlay, and Platform A NIR capture of seated subject indoor dark.}
    \label{fig:data_views}
\end{figure}This protocol introduces deliberate intra-trial variation in respiratory dynamics and provides a controlled way to stress-test the SQI-based rejection and aggregation stages of the pipeline. Ground-truth respiratory rate was approximated from the predefined phases of the controlled breathing protocol rather than from a concurrently worn reference sensor. Accordingly, the expected BPM ranges associated with each breathing phase are used for qualitative validation of the estimated respiratory patterns and rates. Figure~\ref{fig:data_views} presents representative experimental configurations. Table~\ref{tab:dataset} summarizes the resulting dataset composition across modalities, platforms, and distances, totaling 96 sessions of 70-second chest-cropped video. Cells containing 0 correspond to infeasible modality–distance combinations whose operational constraints are analyzed in Section~\ref{sec:system_analysis}. The dataset and associated code will be made available upon request for research purposes, subject to institutional data sharing agreements.

\begin{table}[!ht]
\centering
\caption{Dataset Composition Across Modalities, Platforms, and Distances.}
\label{tab:dataset}
\footnotesize
\begin{tabular}{|l|c|c|c|c|c|}
\hline
\multirow{2}{*}{\textbf{Modality / Platform}} & \multicolumn{4}{c|}{\textbf{Sessions per Distance}} & \multirow{2}{*}{\textbf{Total}} \\
\cline{2-5}
 & \textbf{2m} & \textbf{4m} & \textbf{6m} & \textbf{8m} & \\
\hline
RGB / A,B,C       & 18 & 12 & 12 & 12 & 54 \\
\hline
NIR / A           & 6  & 4  & 4  & 0  & 14 \\
\hline
Thermal / A       & 6  & 4  & 0  & 0  & 10 \\
\hline
Low-Light / B     & 6  & 4  & 4  & 4  & 18 \\
\hline
\multicolumn{5}{|r|}{\textbf{Total sessions}} & \textbf{96} \\
\hline
\end{tabular}
\vspace{4pt}
\par
\begin{minipage}{\columnwidth}
\scriptsize
\raggedright 
\end{minipage}
\end{table}
\section{System Analysis}
\label{sec:system_analysis}

\subsection{Modality-Specific SQI Validity Accuracy}
\begin{figure*}[!t]
  \centering  \includegraphics[width=\textwidth]{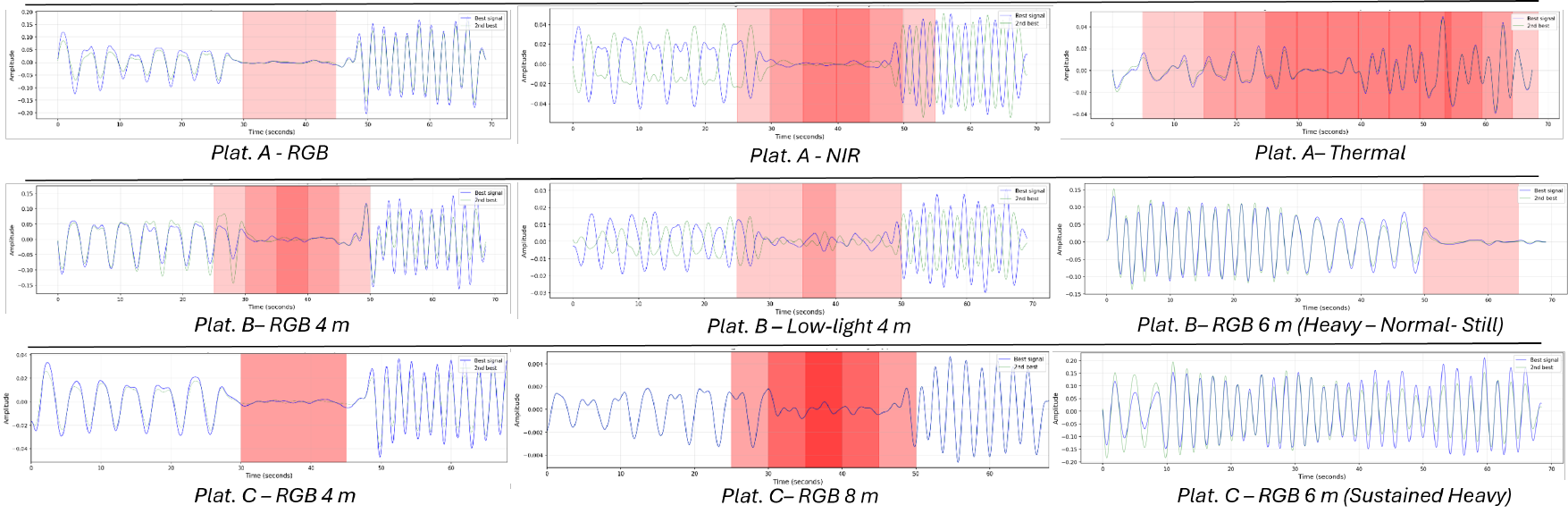}
  \caption{Bandpass filtered respiratory signals (8--35 BPM) across platforms and modalities. Red-shaded regions indicate SQI-rejected windows. Panels at 4\,m show the breath-hold protocol (30--50\,s), while Platform~B and Platform~C panels at 6\,m demonstrate framework response to respiratory pattern variation.}
  \label{fig:exp_signals}
\end{figure*}

Figure~\ref{fig:exp_signals} illustrates representative signal behavior across platforms and sensing modalities. The red-shaded regions denote windows rejected by the proposed SQI framework. \begin{table}[!ht]
\centering
\caption{Modality-Specific SQI Validity and Estimated RR Across Distances.}
\label{tab:sqi_validity}
\footnotesize
\begin{tabular}{|l|c|c|c|c|c|}
\hline
\multirow{2}{*}{\textbf{Modality / Platform}} & \multicolumn{4}{c|}{\textbf{Valid Win. \%}} & \multirow{2}{*}{\textbf{Est. BPM}} \\
\cline{2-5}
 & \textbf{2m} & \textbf{4m} & \textbf{6m} & \textbf{8m} & \\
\hline
RGB / A             & 75   & 92   & 83   & 75   & 16.5 \\
\hline
RGB / B             & 92   & 92   & 75   & 75   & 16.3 \\
\hline
RGB / C             & 83   & 92   & 75   & 75   & 16.0 \\
\hline
NIR / A             & 83   & 67   & 75   & 0.0  & 15.9 \\
\hline
Thermal / A         & 50   & 0.0  & 0.0  & 0.0  & 19.5 \\
\hline
Low-Light / B       & 92.3 & 76.9 & 92.3 & 92.3 & 16.1 \\
\hline
\end{tabular}
% \vspace{0.05in}
% \raggedright
% \scriptsize
\end{table}In particular, the 30--50\,s breath-hold phase in the 4\,m panels is consistently rejected, indicating that the quality index responds to respiratory periodicity and physiological state rather than signal amplitude alone. The Platform~B and Platform~C panels at 6\,m further demonstrate cross-pattern generalization without modality-specific retuning: Platform~B rejects only the still phase in a varying respiratory sequence, whereas Platform~C retains all windows during sustained periodic breathing because they satisfy the SQI criteria.

Table~\ref{tab:sqi_validity} reports the percentage of valid sliding windows and the aggregated respiratory-rate estimate for each modality--platform combination across distance conditions. Because the valid-window percentage is derived entirely from the framework’s internal quality-gating mechanism and does not depend on external ground truth, it provides a reproducible and self-consistent measure of modality reliability that directly reflects the effectiveness of the proposed SQI-based filtering strategy. Across all visible-light modalities, including RGB on Platforms~A, B, and C and the low-light sensor on Platform~B, the framework maintains valid-window rates above 75\% throughout the operational range. RGB on Platform~B shows the most consistent performance, achieving 92\% valid windows at both 2\,m and 4\,m. The corresponding RR estimates remain tightly grouped within 15.9--16.5\,BPM, which is consistent with the controlled breathing protocol used during data collection. NIR on Platform~A performs comparably at short and moderate range but degrades sharply at 8\,m, where no valid windows remain. Thermal imaging on Platform~A yields only marginal validity at 2\,m and complete rejection beyond that distance.

\subsection{Operational Distance Boundaries}
Table~\ref{tab:distance_bounds} summarizes modality-specific operational boundaries across distances and subject poses. RGB provides the broadest operating envelope, supporting reliable RR estimation across all three platforms up to 8\,m under both indoor and outdoor illuminated conditions. \begin{table}[!ht]
\centering
\caption{Modality-Specific Operational Distance and Pose Coverage.}
\label{tab:distance_bounds}
\footnotesize
\begin{tabular}{|l|c|c|c|c|c|c|}
\hline
\textbf{Modality} & \textbf{Platform} & \textbf{2m} & \textbf{4m} & \textbf{6m} & \textbf{8m} & \textbf{Poses} \\
\hline
RGB & A, B, C & $\checkmark$ & $\checkmark$ & $\checkmark$ & $\bullet$ & Stand, Sit, Lie \\
\hline
NIR & A & $\checkmark$ & $\checkmark$ & $\checkmark$ & $\times$ & Stand, Sit, Lie \\
\hline
Thermal & A & $\checkmark$ & $\times$ & $\times$ & $\times$ & Stand, Sit \\
\hline
Low-Light & B & $\checkmark$ & $\checkmark$ & $\checkmark$ & $\bullet$ & Stand, Sit, Lie \\
\hline
\end{tabular}
\end{table} NIR on Platform~A remains effective up to approximately 6\,m, after which performance degrades because of illumination falloff from the fixed-power LED array. This limitation is inherent to the active-illumination design, as emitted intensity decreases with increasing distance. Extending this range would require a higher-power or denser LED configuration, at the expense of additional payload weight and power consumption. Thermal imaging supports valid RR estimation only at 2\,m under dark indoor and outdoor conditions. Beyond this distance, the limited chest-ROI pixel area at the native FLIR~C5 resolution reduces the reliability of pose-keypoint detection, which in turn prevents stable ROI extraction and causes all windows to be rejected by the SQI stage.
The low-light camera on Platform~B performs reliably in pitch-dark environments; however, lying-pose estimation fails beyond 6\,m. We attribute this behavior to a viewing geometry limitation rather than to the sensing modality itself.
Because the camera is mounted near payload height (\textasciitilde 0.6\,m), a lying subject appears with a strongly foreshortened torso profile, reducing the pixel separation between shoulder and hip keypoints. This leads to low-confidence pose estimation and misaligned chest ROI extraction. The same issue is likely to arise on other ground-level robotic platforms with similar camera mounting geometry. 
In Table~\ref{tab:distance_bounds}, entries marked $\times$ denote modality--distance combinations in which the framework fails at both the pose-detection and SQI-validation stages. Entries marked $\bullet$ indicate a different failure mode: pose detection remains successful for standing and seated subjects at 8\,m, but lying-pose keypoint alignment fails because of the viewing-angle constraint, preventing a valid chest ROI from reaching the SQI analysis stage.

\subsection{Platform Computational Performance}

\begin{figure}[!ht]
    \centering
    % \vspace{-0.25in}
    \includegraphics[width=0.45\columnwidth]{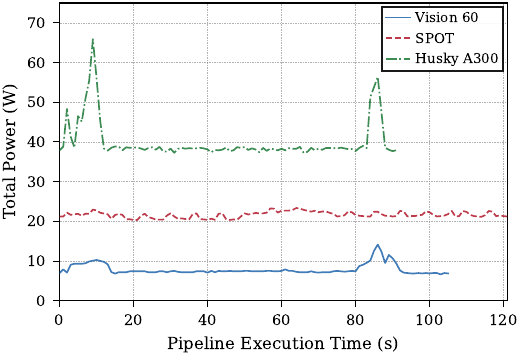}%
    \hspace{0.01\columnwidth}%
    \includegraphics[width=0.45\columnwidth]{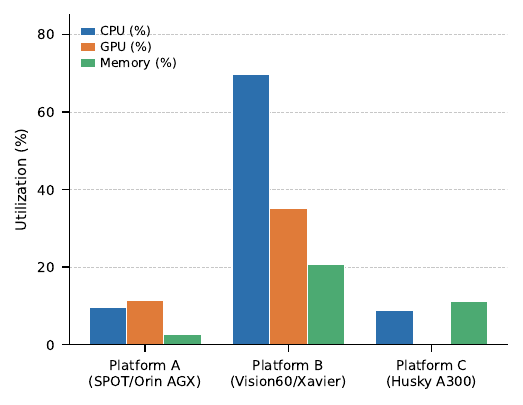}\\
    \vspace{-0.08in}
    {\small (a)} \hspace{0.38\columnwidth} {\small (b)}
    \caption{Platform computational performance during RR pipeline execution: \textit{(a)} total power consumption over time (W); \textit{(b)} average CPU, GPU, and memory utilization.}
    \label{fig:platform_perf}
\end{figure}

Figure~\ref{fig:platform_perf} summarizes the computational characteristics of the full RR monitoring pipeline across the three heterogeneous platforms. Platform~B exhibits the highest resource utilization, with 69.8\% CPU usage and 35\% GPU load, and requires the longest execution time of 2\,min\,20\,s. This behavior reflects the tighter computational constraints of the Jetson Xavier architecture under end-to-end pipeline execution.

By contrast, Platform~A maintains substantially lower utilization, with 9.59\% CPU usage and 11.53\% GPU load, while sustaining a relatively stable average power draw of approximately 22\,W over a 2-minute run. This improved efficiency is attributable to the higher compute density and core capacity of the Jetson Orin AGX. Platform~C achieves the shortest execution time, 1\,min\,34\,s, but does so with considerably higher peak power consumption in the 38--65\,W range, consistent with its x86 host and desktop-class RTX~4060~Ti GPU. Overall, all three platforms support real-time execution of the proposed pipeline within practical field-deployment constraints.

\section{Discussion}
The results demonstrate both the practical value and the current limits of the proposed modality-adaptive contactless RR monitoring framework across heterogeneous robotic platforms. The consistently high valid-window percentages observed under visible-light sensing, ranging from 75\% to 92\% across Platforms~A, B, and C, show that keypoint-guided chest ROI extraction and SQI-based filtering generalize across multiple robots and edge-computing architectures without platform-specific redesign. This cross-platform consistency is important for search-and-rescue and disaster-response settings, where robotic systems may differ in mobility, sensing payload, and onboard compute capability, yet still need to support a common triage workflow~\cite{darpaTriagChallenge,tranzatto2022team}. The experiments also reveal several sensing and geometry constraints. Thermal sensing is currently limited by the native resolution of the FLIR~C5, which restricts reliable operation to short range. Similarly, the lying-pose failures beyond 6\,m for ground-mounted cameras arise primarily from viewing geometry rather than sensing modality: a low camera height causes torso foreshortening and reduces shoulder--hip keypoint separation, making ROI extraction less reliable. These findings show that contactless respiratory monitoring on mobile robots depends not only on sensing modality, but also on the interaction among sensor placement, subject posture, and stand-off distance. In addition, the current framework operates on fixed video segments while the robot remains stationary. Although appropriate for controlled evaluation, practical field deployment will require continuous monitoring during robot motion, where vibration, viewpoint change, and ROI drift must be addressed. Overall, the results suggest that SQI-gated multimodal sensing provides a strong basis for robotic contactless physiological monitoring, while also highlighting clear directions for system refinement.

\section{Conclusion}
We presented a modality-adaptive contactless respiratory monitoring framework for three heterogeneous mobile robots with different locomotion, sensing, and edge-computing configurations. By combining brightness-adaptive sensor selection, pose-guided chest ROI extraction, and SQI-based window filtering, the framework supports unified RR monitoring across RGB, thermal, NIR, and low-light modalities without per-platform retuning. The results establish clear operational envelopes, with RGB performing up to 8\,m across all platforms, NIR up to 6\,m, thermal at 2\,m, and low-light supporting monitoring up to 8\,m in complete darkness. Future work will focus on extending the thermal operating range with higher-resolution thermal sensors, improving long-range lying-pose monitoring through better camera geometry or adaptive robot positioning, and enabling continuous RR estimation during robot locomotion. We also plan to incorporate additional contactless vital signs, such as heart rate, to support broader multi-vital robotic assessment in hazardous environments.

\section*{Acknowledgement}
This work has been partially supported by NSF CNS EAGER Grant \#2233879, NSF IIS Grant \#2509680, NSF CAREER Grant \#1750936, NSF I-Corps Grant \# 2502886, U.S. Army Grants \#W911NF2120076 \& \#W911NF2410367, and ONR Grant \#N00014-23-1-2119.

\bibliographystyle{IEEEtran}
\bibliography{references}

\end{document}